\documentclass[11pt]{article}

\usepackage[utf8]{inputenc}
\usepackage[T2A,T1]{fontenc}
\usepackage[ukrainian,english]{babel}
\usepackage{amsmath,amssymb}
\newcommand{\ukr}[1]{\foreignlanguage{ukrainian}{#1}}
\usepackage{tikz}
\usepackage{booktabs}
\usepackage{multirow}
\usepackage{graphicx}
\graphicspath{{figures/}}
\usepackage{caption}
\usepackage{subcaption}
\usepackage{hyperref}
\usepackage[numbers,sort&compress]{natbib}
\usepackage{geometry}
\title{Temporal Concept Drift in Legal Judgment Prediction:\\
Neural Baselines Across Three Epochs of Ukrainian Court Decisions}

\author{Volodymyr Ovcharov\\
\texttt{overthelex@legal.org.ua}}

\date{}

\begin{document}

\maketitle

\begin{abstract}
Legal NLP benchmarks evaluate models on randomly split data, implicitly assuming that legal language is stationary.
We test this assumption by fine-tuning four transformer encoders -- XLM-RoBERTa (base and large) and their legal-domain variants -- on Ukrainian court decisions from three temporal epochs defined by geopolitical disruptions: pre-war (2008--2013), hybrid war (2014--2021), and full-scale invasion (2022--2026).
Each model is trained on one epoch and evaluated on all three, producing a $3 \times 3$ cross-temporal generalization matrix.

Four findings emerge.
\textbf{(1)} Forward degradation is severe: models trained on pre-war data lose up to 27.2 percentage points of macro-F1 when applied to full-scale invasion era decisions, confirming and extending the 27.9-pp gap observed with classical baselines~\cite{ovcharov2025tokenizer}.
\textbf{(2)} The degradation is asymmetric: backward transfer (full-scale$\to$pre-war) is substantially more robust than forward transfer, consistent with the hypothesis that legal language is additive -- new legal frameworks subsume older ones, but the reverse does not hold.
\textbf{(3)} Legal-domain pretraining (Legal-XLM-R) does not improve absolute performance compared to general-purpose XLM-R, but reduces forward degradation magnitude and asymmetry, suggesting that domain pretraining captures more temporally stable -- if less discriminative -- representations.
\textbf{(4)} Chronological continual learning (sequential fine-tuning pre-war$\to$hybrid$\to$full-scale) eliminates catastrophic forgetting for general XLM-R: pre-war knowledge is fully retained (+1.8 to +6.2 pp) while full-scale performance gains +16.5 to +19.0 pp. Reverse-chronological continual learning, however, causes severe forgetting ($-$12.2 to $-$14.3 pp on full-scale), and Legal-XLM-R forgets in both directions. This directional asymmetry in continual learning reinforces the additive-language hypothesis from a complementary angle.

Cross-jurisdictional pretraining on Swiss Judgment Prediction data improves absolute performance (+3 to +10~pp) but does not reduce temporal degradation magnitude (20.3 vs.\ 21.3~pp forward gap), confirming that temporal drift is an intrinsic property of legal language evolution, not a jurisdiction-specific artifact.

These results establish the first neural temporal robustness benchmark for legal NLP and demonstrate that temporal drift is a dominant, underexplored source of performance degradation -- exceeding the impact of model selection, domain pretraining, and cross-jurisdictional transfer.
Chronological retraining is shown to be an effective mitigation strategy for general-purpose models.
The dataset (428K decisions across three epochs) is publicly available as a LEXTREME contribution.
\end{abstract}

\section{Introduction}

The performance of NLP models degrades when the temporal distribution of test data diverges from training data -- a phenomenon known as temporal concept drift~\cite{lazaridou2021temporal,luu2022temporal}.
In general NLP, this manifests as outdated factual knowledge and shifting linguistic conventions.
In the legal domain, the effect is amplified by three structural factors: legislative change introduces new statutes and modifies existing ones; judicial practice evolves as courts interpret new legislation; and exogenous shocks (reforms, conflicts) can alter the entire procedural framework within which courts operate.

Despite this, the dominant legal NLP benchmarks -- LexGLUE~\cite{chalkidis2021lexglue}, LEXTREME~\cite{niklaus2023lextreme}, and SCALE~\cite{niklaus2023scale} -- evaluate models on randomly split data, treating temporal variation as noise rather than signal.
This design choice is understandable for benchmarks focused on cross-lingual comparison, but it obscures a critical question for practitioners: \emph{how quickly does a deployed legal NLP model become unreliable?}

We address this question using a natural experiment.
Ukraine's judicial system operated under three distinct regimes over the period 2008--2026:
\begin{enumerate}
  \item \textbf{Pre-war (2008--2013):} Peacetime baseline. All 832 courts operational, stable procedural rules.
  \item \textbf{Hybrid war (2014--2021):} Crimea annexation (2014), loss of courts in occupied territories, judicial reform (2017), procedural modernization.
  \item \textbf{Full-scale invasion (2022--2026):} Martial law, altered procedural timelines, new Criminal Code articles (collaborationism, aiding the aggressor state), surge in military criminal cases.
\end{enumerate}
These three epochs are not arbitrary periodizations -- they correspond to structural breaks in citation network topology~\cite{ovcharov2025citation}, 33--47\% decay in co-citation predictability~\cite{ovcharov2025statute}, and 27.9-pp degradation in classical (TF-IDF) judgment prediction~\cite{ovcharov2025tokenizer}.
The present work extends these findings to neural models, answering the question: \emph{does fine-tuned transformer performance degrade in the same way, and can legal-domain pretraining mitigate the effect?}

\paragraph{Contributions.}
\begin{enumerate}
  \item We produce the first \textbf{neural cross-temporal generalization matrix} for legal judgment prediction, fine-tuning four XLM-R variants on 428K Ukrainian court decisions across three epochs.
  \item We quantify \textbf{forward--backward asymmetry} in neural temporal transfer and test whether legal-domain pretraining provides temporal robustness.
  \item We evaluate \textbf{continual learning} across temporal epochs, showing that chronological retraining eliminates catastrophic forgetting for general models while reverse-chronological retraining causes severe forgetting -- a directional asymmetry that reinforces the additive-language hypothesis.
  \item We conduct \textbf{cross-jurisdictional temporal transfer} experiments using Swiss Judgment Prediction~\cite{niklaus2021swiss}, extending Cross-X transfer~\cite{niklaus2022cross} to the temporal dimension and showing that foreign-jurisdiction pretraining does not mitigate temporal degradation.
  \item We release the \textbf{dataset} (428K decisions, three epochs, chronological splits) as a LEXTREME contribution -- the first Cyrillic-script subset with temporal annotations.
\end{enumerate}

\section{Related Work}

\paragraph{Legal NLP benchmarks.}
LexGLUE~\cite{chalkidis2021lexglue} established a multi-task benchmark for English legal NLU, including case law from the European Court of Human Rights.
LEXTREME~\cite{niklaus2023lextreme} extended this to 11 datasets across 24 EU languages, evaluating XLM-R and domain-specific legal models.
Both benchmarks use random train/test splits.
SCALE~\cite{niklaus2023scale} introduced longer-document tasks from the Swiss legal system, and FairLex~\cite{chalkidis2022fairlex} added a fairness dimension.
None of these benchmarks control for temporal distribution shift.

\paragraph{Legal judgment prediction.}
Swiss Judgment Prediction~\cite{niklaus2021swiss} provided the first multilingual legal judgment prediction benchmark, with decisions spanning 2000--2020 but evaluated on random splits.
PILOT~\cite{cao2024pilot} introduced temporal pattern handling for case law retrieval, but focused on precedent identification rather than judgment prediction.
Cross-X transfer~\cite{niklaus2022cross} examined cross-lingual, cross-domain, and cross-regional transfer in legal NLP, but not cross-temporal transfer.
We extend Cross-X to the temporal dimension, testing whether cross-jurisdictional pretraining mitigates temporal degradation.

\paragraph{Legal-domain language models.}
LEGAL-BERT~\cite{chalkidis2020legal} demonstrated the value of domain-specific pretraining for English legal text.
The MultiLegalPile corpus~\cite{niklaus2023multilegalpile} enabled pretraining of multilingual legal models (Legal-XLM-R), which achieved state-of-the-art on LEXTREME.
SaulLM~\cite{colombo2024saullm} scaled legal domain adaptation to 54B and 141B parameters.
LeXFiles~\cite{chalkidis2023lexfiles} provided a multinational English legal corpus with probing tasks.
LEMUR~\cite{ahmadi2026lemur} introduced multilingual legal embedding models for retrieval.
A key question we address is whether legal-domain pretraining captures temporally stable representations.

\paragraph{Temporal generalization in NLP.}
Lazaridou et al.~\cite{lazaridou2021temporal} demonstrated that language models degrade on temporally shifted data, with performance inversely proportional to temporal distance.
Luu et al.~\cite{luu2022temporal} showed that temporal misalignment affects multiple NLP tasks beyond language modeling.
Dhingra et al.~\cite{dhingra2022time} proposed time-aware language models to mitigate temporal degradation.
In the legal domain, Ovcharov~\cite{ovcharov2025tokenizer} established a 27.9-pp forward degradation gap using TF-IDF classifiers on Ukrainian court decisions, but explicitly noted the absence of neural baselines as a limitation.

\paragraph{Continual learning.}
Catastrophic forgetting~\cite{kirkpatrick2017overcoming} is a fundamental challenge when models are sequentially trained on new data distributions.
In the legal domain, temporal epochs form a natural curriculum: each period introduces new legislation and procedural frameworks that build on prior ones.
We evaluate whether this additive structure enables sequential fine-tuning without catastrophic forgetting, testing both chronological (forward) and reverse-chronological (backward) training orders.

\section{Dataset}

\subsection{Source and Extraction}

We extract court decisions from the Unified State Register of Court Decisions (EDRSR, \ukr{ЄДРСР}), a publicly accessible database of all Ukrainian court decisions since 2006.
The register contains over 100 million documents.
We focus on civil and commercial jurisdictions, which provide the most consistent case structure across temporal epochs.

Each document is processed as follows: (1)~the facts section (\ukr{встановив}) is extracted as model input; (2)~the dispositive section (\ukr{вирішив}) is parsed for outcome classification; (3)~personally identifiable information is replaced with placeholder tokens (\texttt{[PERSON]}, \texttt{[ADDRESS]}, \texttt{[NUMBER]}).
Texts are truncated to 10,000 characters.

\subsection{Temporal Epochs}

Decisions are divided into three epochs reflecting major geopolitical disruptions:
\begin{itemize}
  \item \textbf{Pre-war (2008--2013):} 128,075 decisions. Peacetime judicial baseline.
  \item \textbf{Hybrid war (2014--2021):} 150,000 decisions. Post-Crimea annexation, judicial reform 2017.
  \item \textbf{Full-scale invasion (2022--2026):} 150,000 decisions. Martial law, procedural changes.
\end{itemize}

\subsection{Label Schema}

Outcomes are classified into three categories via regex extraction from the dispositive section:
\begin{itemize}
  \item \textbf{Approved} (\ukr{задоволено}): claim fully granted.
  \item \textbf{Dismissed} (\ukr{відмовлено}): claim denied.
  \item \textbf{Partial} (\ukr{частково задоволено}): claim partially granted.
\end{itemize}
The pre-war epoch has a slight class imbalance (50K approved / 28K dismissed / 50K partial); the other two epochs are balanced at 50K per class.
This imbalance reflects the natural distribution of outcomes in pre-war civil litigation.

\subsection{Chronological Splits}

Within each epoch, documents are split chronologically: the earliest 80\% form the training set, the next 10\% the validation set, and the most recent 10\% the test set.
This prevents temporal leakage within epochs and ensures that models are always evaluated on decisions that postdate their training data.

\begin{table}[t]
\centering
\caption{Dataset statistics. All splits are chronological within each epoch.}
\label{tab:dataset}
\begin{tabular}{lcccc}
\toprule
\textbf{Epoch} & \textbf{Period} & \textbf{Train} & \textbf{Val} & \textbf{Test} \\
\midrule
Pre-war     & 2008--2013 & 102,460 & 12,807 & 12,808 \\
Hybrid war  & 2014--2021 & 120,000 & 15,000 & 15,000 \\
Full-scale  & 2022--2026 & 120,000 & 15,000 & 15,000 \\
\midrule
\textbf{Total} & 2008--2026 & 342,460 & 42,807 & 42,808 \\
\bottomrule
\end{tabular}
\end{table}

\section{Experimental Setup}

\subsection{Models}

We evaluate four XLM-RoBERTa~\cite{conneau2020xlmr} variants, covering the interaction of model scale and domain pretraining:
\begin{enumerate}
  \item \textbf{XLM-R Base} (278M parameters) -- general multilingual baseline.
  \item \textbf{XLM-R Large} (560M parameters) -- scale comparison.
  \item \textbf{Legal-XLM-R Base} (278M) -- pretrained on the 689GB MultiLegalPile~\cite{niklaus2023multilegalpile}.
  \item \textbf{Legal-XLM-R Large} (560M) -- legal-domain pretrained + scale.
\end{enumerate}
This $2 \times 2$ design (general vs.\ legal, base vs.\ large) isolates the effects of domain pretraining and model capacity on temporal robustness.

\subsection{Training Configuration}

All models are fine-tuned with the HuggingFace Transformers library~\cite{wolf2020transformers} using the AdamW optimizer~\cite{loshchilov2019decoupled}.
Training configuration:
\begin{itemize}
  \item Learning rate: $2 \times 10^{-5}$ (base), $1 \times 10^{-5}$ (large)
  \item Weight decay: 0.01
  \item Maximum sequence length: 512 tokens
  \item Training epochs: 5, with early stopping (patience 2) on validation macro-F1
  \item Warmup: 10\% of total steps
  \item Batch size: 16 per GPU (base), 8 per GPU (large), with gradient accumulation
  \item Hardware: NVIDIA A10G GPUs (AWS ml.g5 instances)
\end{itemize}
Each experiment is run with three random seeds (42, 123, 456) and results are reported as mean $\pm$ standard deviation.

\subsection{Evaluation Protocol}

\paragraph{Experiment 1: In-epoch baselines.}
Each model is trained on epoch $E_i$ and evaluated on the test split of the same epoch.
This establishes the diagonal of the generalization matrix: the best-case performance when train and test distributions match.

\paragraph{Experiment 2: Cross-epoch generalization.}
Each of the 12 trained models (4 models $\times$ 3 epochs) is evaluated on all three test splits, producing a $3 \times 3$ macro-F1 matrix per model.
Key derived metrics:
\begin{itemize}
  \item \textbf{Forward degradation:} $\Delta_\text{fwd} = F1(E_1 \to E_1) - F1(E_1 \to E_3)$
  \item \textbf{Backward degradation:} $\Delta_\text{bwd} = F1(E_3 \to E_3) - F1(E_3 \to E_1)$
  \item \textbf{Asymmetry gap:} $\Delta_\text{fwd} - \Delta_\text{bwd}$
\end{itemize}

\paragraph{Experiment 3: Cross-jurisdictional temporal transfer.}
We use Swiss Judgment Prediction (SJP)~\cite{niklaus2021swiss} as a foreign-jurisdiction source, extending the Cross-X transfer framework~\cite{niklaus2022cross} to the temporal dimension.
SJP contains multilingual (German, French, Italian) Swiss Federal Supreme Court decisions (2000--2020) with binary outcome labels (approval/dismissal).
We evaluate three cross-jurisdictional settings with XLM-R Base:
\begin{itemize}
  \item \textbf{Zero-shot:} Train on SJP, test on Ukrainian (binary, dropping partial class).
  \item \textbf{Transfer:} Phase~1: fine-tune on SJP (binary). Phase~2: fine-tune on Ukrainian pre-war (3-class). Test on all three Ukrainian epochs.
  \item \textbf{Reverse:} Train on Ukrainian hybrid-war (binary), test on SJP by language.
\end{itemize}
The transfer setting produces 3-class predictions on Ukrainian data, enabling direct comparison with Table~\ref{tab:cross_epoch}.
Each setting is run with three seeds (42, 123, 456).

\paragraph{Experiment 4: Continual learning.}
Each model is fine-tuned sequentially across all three epochs in two directions:
\begin{itemize}
  \item \textbf{Forward (chronological):} pre-war $\to$ hybrid $\to$ full-scale.
  \item \textbf{Backward (reverse-chronological):} full-scale $\to$ hybrid $\to$ pre-war.
\end{itemize}
After each stage, the model is evaluated on all three test splits.
This produces a trajectory of macro-F1 values across stages, revealing whether knowledge from earlier stages is retained or forgotten.
Each direction is run with available seeds (2 seeds for forward, 1--2 for backward per model; see Table~\ref{tab:cl_summary}).
Training uses the same hyperparameters as Experiments~1--2.

\paragraph{Metrics.}
We report macro-F1 as the primary metric (robust to class imbalance), along with per-class F1 and accuracy.
Statistical significance is assessed via paired bootstrap with 1,000 iterations.

\section{Results}

\subsection{In-Epoch Baselines}

Table~\ref{tab:in_epoch} presents in-epoch performance for all four models alongside the TF-IDF baseline from~\cite{ovcharov2025tokenizer}.

\begin{table}[t]
\centering
\caption{In-epoch performance (diagonal of generalization matrix). Macro-F1 (\%) averaged over 3 seeds.}
\label{tab:in_epoch}
\begin{tabular}{lcccc}
\toprule
\textbf{Model} & \textbf{Pre-war} & \textbf{Hybrid} & \textbf{Full-scale} & \textbf{Mean} \\
\midrule
XLM-R Base          & 64.4{\small$\pm$0.6} & 66.9{\small$\pm$1.1} & 59.1{\small$\pm$0.8} & 63.4 \\
XLM-R Large         & 67.1{\small$\pm$1.8} & 68.9{\small$\pm$0.4} & 58.1{\small$\pm$0.2} & 64.7 \\
Legal-XLM-R Base    & 52.2{\small$\pm$6.5} & 60.7{\small$\pm$1.0} & 54.9{\small$\pm$0.8} & 55.9 \\
Legal-XLM-R Large   & 56.4{\small$\pm$2.4} & 61.3{\small$\pm$0.8} & 54.3{\small$\pm$0.6} & 57.4 \\
\midrule
TF-IDF (baseline)   & 86.5 & 83.7 & 69.3 & 79.8 \\
\bottomrule
\end{tabular}
\end{table}

Two patterns are immediately apparent.
First, all neural models substantially underperform the TF-IDF baseline -- by 15--30 percentage points depending on epoch and model.
This is consistent with the finding of Ovcharov~\cite{ovcharov2025tokenizer} that tokenizer fertility penalties on Ukrainian Cyrillic text severely limit transformer effectiveness: XLM-R's SentencePiece tokenizer fragments Ukrainian legal terms into 3--5 subwords, reducing the effective context window and forcing models to reconstruct word-level semantics from fragmentary representations.

Second, all models achieve their lowest scores on the full-scale epoch, confirming that post-invasion decisions present a harder classification target.
This is expected: martial law introduced novel procedural rules, new criminal categories, and disrupted the court system's geographic coverage, creating a distribution shift even within the same task.

General-purpose XLM-R outperforms its legal-domain counterpart by 7--9 pp on average, with the gap widest on the pre-war epoch (12.2 pp for base, 10.7 pp for large).
The high variance of Legal-XLM-R Base on pre-war data ($\pm$6.5) suggests unstable convergence, possibly due to conflicting signals between the legal pretraining corpus and the Ukrainian-specific fine-tuning data.

\subsection{Cross-Epoch Generalization}

\begin{figure*}[t]
\centering
\input{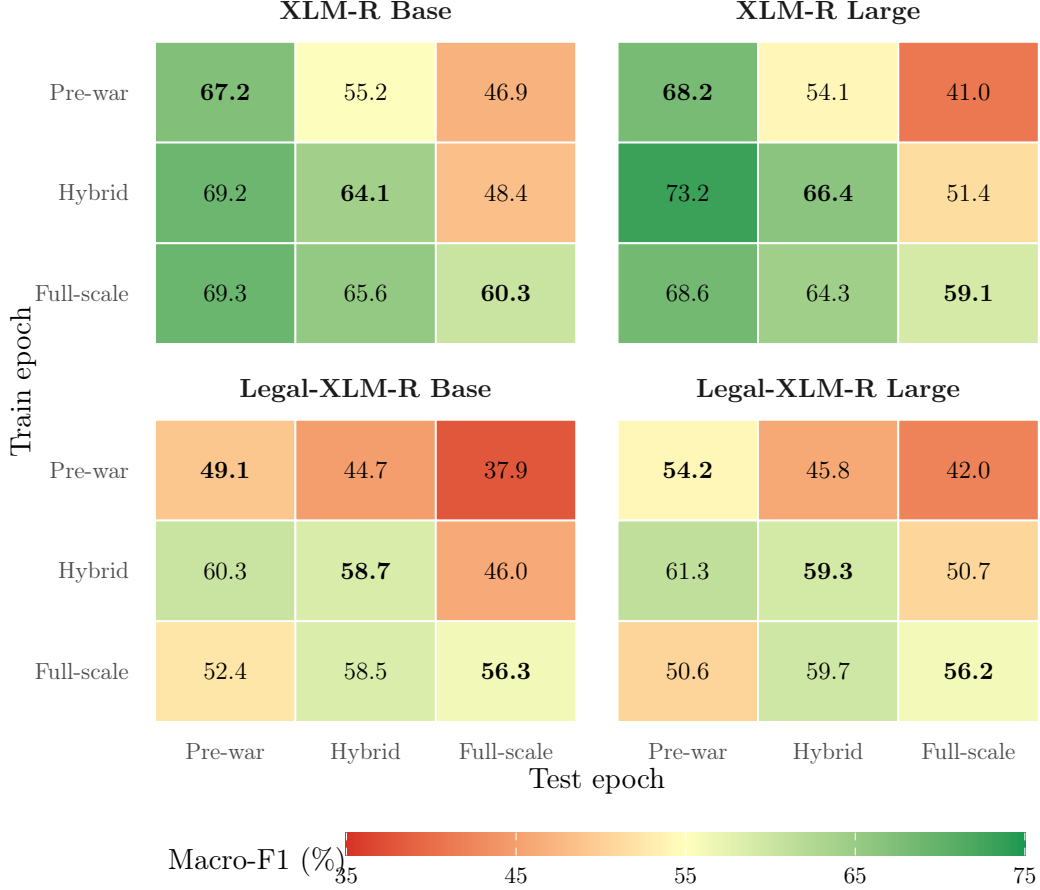}
\caption{Cross-epoch generalization matrices for all four models. Rows: training epoch; columns: test epoch. Diagonal cells (in-epoch) are consistently the highest per column for general models, but off-diagonal backward transfer sometimes exceeds the diagonal. The full-scale column is universally the hardest.}
\label{fig:heatmap}
\end{figure*}

Table~\ref{tab:cross_epoch} and Figure~\ref{fig:heatmap} present the full $3 \times 3$ cross-epoch generalization matrices for all four models.

\begin{table*}[t]
\centering
\caption{Cross-epoch generalization matrices. Macro-F1 (\%) averaged over 3 seeds. Rows: training epoch; columns: test epoch. \textbf{Bold}: in-epoch (diagonal). Best off-diagonal per column underlined.}
\label{tab:cross_epoch}
\begin{tabular}{ll ccc}
\toprule
\textbf{Model} & \textbf{Train $\backslash$ Test} & \textbf{Pre-war} & \textbf{Hybrid} & \textbf{Full-scale} \\
\midrule
\multirow{3}{*}{XLM-R Base}
  & Pre-war    & \textbf{67.2}{\small$\pm$0.8} & 55.2{\small$\pm$1.1} & 46.9{\small$\pm$4.1} \\
  & Hybrid     & \underline{69.2}{\small$\pm$2.2} & \textbf{64.1}{\small$\pm$1.2} & 48.4{\small$\pm$2.1} \\
  & Full-scale & 69.3{\small$\pm$1.8} & \underline{65.6}{\small$\pm$1.2} & \textbf{60.3}{\small$\pm$0.6} \\
\midrule
\multirow{3}{*}{XLM-R Large}
  & Pre-war    & \textbf{68.2}{\small$\pm$1.7} & 54.1{\small$\pm$0.7} & 41.0{\small$\pm$1.3} \\
  & Hybrid     & \underline{73.2}{\small$\pm$0.3} & \textbf{66.4}{\small$\pm$0.5} & \underline{51.4}{\small$\pm$3.1} \\
  & Full-scale & 68.6{\small$\pm$1.8} & \underline{64.3}{\small$\pm$0.9} & \textbf{59.1}{\small$\pm$0.7} \\
\midrule
\multirow{3}{*}{Legal-XLM-R Base}
  & Pre-war    & \textbf{49.1}{\small$\pm$8.1} & 44.7{\small$\pm$3.6} & 37.9{\small$\pm$2.9} \\
  & Hybrid     & \underline{60.3}{\small$\pm$1.3} & \textbf{58.7}{\small$\pm$0.9} & 46.0{\small$\pm$3.1} \\
  & Full-scale & 52.4{\small$\pm$5.3} & \underline{58.5}{\small$\pm$0.6} & \textbf{56.3}{\small$\pm$0.6} \\
\midrule
\multirow{3}{*}{Legal-XLM-R Large}
  & Pre-war    & \textbf{54.2}{\small$\pm$4.2} & 45.8{\small$\pm$1.9} & 42.0{\small$\pm$2.9} \\
  & Hybrid     & \underline{61.3}{\small$\pm$0.3} & \textbf{59.3}{\small$\pm$0.8} & \underline{50.7}{\small$\pm$1.2} \\
  & Full-scale & 50.6{\small$\pm$2.8} & \underline{59.7}{\small$\pm$0.5} & \textbf{56.2}{\small$\pm$0.8} \\
\bottomrule
\end{tabular}
\end{table*}

Several patterns emerge from the cross-epoch matrices.

\textit{Forward degradation is severe across all models.}
Pre-war trained models lose 11.2--27.2 pp when evaluated on full-scale data, with larger models degrading more: XLM-R Large loses 27.2 pp (68.2 $\to$ 41.0), compared to 20.3 pp for XLM-R Base (67.2 $\to$ 46.9).
This suggests that larger models overfit more to epoch-specific distributional features.

\textit{The full-scale epoch is universally hardest.}
Regardless of training epoch, all models achieve their lowest scores when tested on full-scale data.
Even hybrid-trained models -- temporally adjacent to the full-scale epoch -- lose 15.7--18.6 pp on full-scale compared to their in-epoch performance.

\textit{Backward transfer is not degradation but improvement for general models.}
Full-scale trained XLM-R Base scores 69.3 on pre-war test data -- higher than its own in-epoch performance of 60.3.
Similarly, XLM-R Large trained on full-scale scores 68.6 on pre-war, compared to 59.1 on its native epoch.
This striking asymmetry suggests that models trained on the more complex full-scale distribution learn representations that generalize well to the simpler pre-war distribution, but not vice versa.

\textit{Hybrid-trained models are the best general-purpose choice.}
Across all four model families, hybrid-trained models achieve the highest or near-highest off-diagonal scores.
XLM-R Large trained on hybrid achieves 73.2 on pre-war (exceeding the pre-war model's own 68.2) and 51.4 on full-scale (the best off-diagonal score for that column).
The hybrid epoch spans the longest period (8 years) and encompasses the most diverse legal landscape, which may explain its superior transfer properties.

\subsection{Legal-Domain Pretraining Effect}

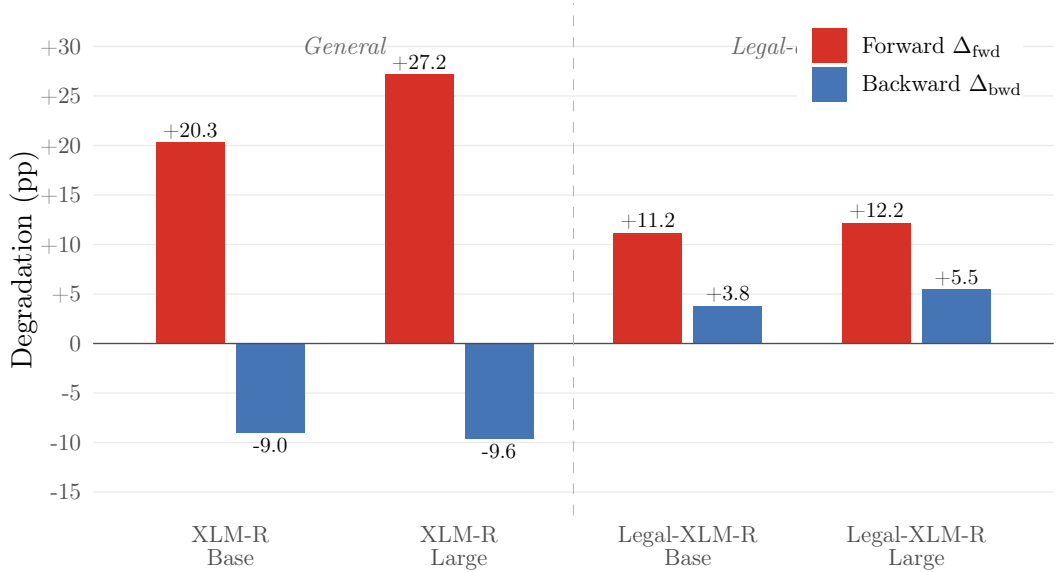
\begin{figure}[t]
\centering
\begin{tikzpicture}[x=1pt,y=1pt]
\definecolor{fillColor}{RGB}{255,255,255}
\path[use as bounding box,fill=fillColor,fill opacity=0.00] (0,0) rectangle (397.48,216.81);
\begin{scope}
\path[clip] (  0.00,  0.00) rectangle (397.48,216.81);
\definecolor{fillColor}{RGB}{255,255,255}

\path[fill=fillColor] ( -0.00,  0.00) rectangle (397.48,216.81);
\end{scope}
\begin{scope}
\path[clip] ( 32.05, 22.20) rectangle (393.48,214.81);
\definecolor{drawColor}{gray}{0.92}

\path[draw=drawColor,line width= 0.5pt,line join=round] ( 32.05, 30.96) --
	(393.48, 30.96);

\path[draw=drawColor,line width= 0.5pt,line join=round] ( 32.05, 49.59) --
	(393.48, 49.59);

\path[draw=drawColor,line width= 0.5pt,line join=round] ( 32.05, 68.21) --
	(393.48, 68.21);

\path[draw=drawColor,line width= 0.5pt,line join=round] ( 32.05, 86.84) --
	(393.48, 86.84);

\path[draw=drawColor,line width= 0.5pt,line join=round] ( 32.05,105.47) --
	(393.48,105.47);

\path[draw=drawColor,line width= 0.5pt,line join=round] ( 32.05,124.10) --
	(393.48,124.10);

\path[draw=drawColor,line width= 0.5pt,line join=round] ( 32.05,142.72) --
	(393.48,142.72);

\path[draw=drawColor,line width= 0.5pt,line join=round] ( 32.05,161.35) --
	(393.48,161.35);

\path[draw=drawColor,line width= 0.5pt,line join=round] ( 32.05,179.98) --
	(393.48,179.98);

\path[draw=drawColor,line width= 0.5pt,line join=round] ( 32.05,198.60) --
	(393.48,198.60);
\definecolor{fillColor}{RGB}{215,48,39}

\path[fill=fillColor] ( 55.72, 86.84) rectangle ( 81.53,162.47);
\definecolor{fillColor}{RGB}{69,117,180}

\path[fill=fillColor] ( 85.83, 53.31) rectangle (111.65, 86.84);
\definecolor{fillColor}{RGB}{215,48,39}

\path[fill=fillColor] (141.77, 86.84) rectangle (167.59,188.17);
\definecolor{fillColor}{RGB}{69,117,180}

\path[fill=fillColor] (171.89, 51.08) rectangle (197.71, 86.84);
\definecolor{fillColor}{RGB}{215,48,39}

\path[fill=fillColor] (227.83, 86.84) rectangle (253.64,128.57);
\definecolor{fillColor}{RGB}{69,117,180}

\path[fill=fillColor] (257.95, 86.84) rectangle (283.76,101.00);
\definecolor{fillColor}{RGB}{215,48,39}

\path[fill=fillColor] (313.88, 86.84) rectangle (339.70,132.29);
\definecolor{fillColor}{RGB}{69,117,180}

\path[fill=fillColor] (344.00, 86.84) rectangle (369.82,107.33);
\definecolor{drawColor}{gray}{0.30}

\path[draw=drawColor,line width= 0.5pt,line join=round] ( 32.05, 86.84) -- (393.48, 86.84);
\definecolor{drawColor}{RGB}{0,0,0}

\node[text=drawColor,anchor=base,inner sep=0pt, outer sep=0pt, scale=  0.74] at ( 68.62,164.51) {+20.3};

\node[text=drawColor,anchor=base,inner sep=0pt, outer sep=0pt, scale=  0.74] at ( 98.74, 46.18) {-9.0};

\node[text=drawColor,anchor=base,inner sep=0pt, outer sep=0pt, scale=  0.74] at (154.68,190.21) {+27.2};

\node[text=drawColor,anchor=base,inner sep=0pt, outer sep=0pt, scale=  0.74] at (184.80, 43.94) {-9.6};

\node[text=drawColor,anchor=base,inner sep=0pt, outer sep=0pt, scale=  0.74] at (240.74,130.60) {+11.2};

\node[text=drawColor,anchor=base,inner sep=0pt, outer sep=0pt, scale=  0.74] at (270.86,103.04) {+3.8};

\node[text=drawColor,anchor=base,inner sep=0pt, outer sep=0pt, scale=  0.74] at (326.79,134.33) {+12.2};

\node[text=drawColor,anchor=base,inner sep=0pt, outer sep=0pt, scale=  0.74] at (356.91,109.37) {+5.5};
\definecolor{drawColor}{gray}{0.40}

\node[text=drawColor,anchor=base,inner sep=0pt, outer sep=0pt, scale=  0.85] at (126.71,195.66) {\textit{General}};

\node[text=drawColor,anchor=base,inner sep=0pt, outer sep=0pt, scale=  0.85] at (298.82,195.66) {\textit{Legal-domain}};
\definecolor{drawColor}{gray}{0.70}

\path[draw=drawColor,line width= 0.3pt,dash pattern=on 4pt off 4pt ,line join=round] (212.77, 22.20) -- (212.77,214.81);
\end{scope}
\begin{scope}
\path[clip] (  0.00,  0.00) rectangle (397.48,216.81);
\definecolor{drawColor}{gray}{0.30}

\node[text=drawColor,anchor=base east,inner sep=0pt, outer sep=0pt, scale=  0.80] at ( 27.55, 28.20) {-15};

\node[text=drawColor,anchor=base east,inner sep=0pt, outer sep=0pt, scale=  0.80] at ( 27.55, 46.83) {-10};

\node[text=drawColor,anchor=base east,inner sep=0pt, outer sep=0pt, scale=  0.80] at ( 27.55, 65.46) {-5};

\node[text=drawColor,anchor=base east,inner sep=0pt, outer sep=0pt, scale=  0.80] at ( 27.55, 84.09) {0};

\node[text=drawColor,anchor=base east,inner sep=0pt, outer sep=0pt, scale=  0.80] at ( 27.55,102.71) {+5};

\node[text=drawColor,anchor=base east,inner sep=0pt, outer sep=0pt, scale=  0.80] at ( 27.55,121.34) {+10};

\node[text=drawColor,anchor=base east,inner sep=0pt, outer sep=0pt, scale=  0.80] at ( 27.55,139.97) {+15};

\node[text=drawColor,anchor=base east,inner sep=0pt, outer sep=0pt, scale=  0.80] at ( 27.55,158.60) {+20};

\node[text=drawColor,anchor=base east,inner sep=0pt, outer sep=0pt, scale=  0.80] at ( 27.55,177.22) {+25};

\node[text=drawColor,anchor=base east,inner sep=0pt, outer sep=0pt, scale=  0.80] at ( 27.55,195.85) {+30};
\end{scope}
\begin{scope}
\path[clip] (  0.00,  0.00) rectangle (397.48,216.81);
\definecolor{drawColor}{gray}{0.30}

\node[text=drawColor,anchor=base,inner sep=0pt, outer sep=0pt, scale=  0.80] at ( 83.68, 12.20) {XLM-R};

\node[text=drawColor,anchor=base,inner sep=0pt, outer sep=0pt, scale=  0.80] at ( 83.68,  3.56) {Base};

\node[text=drawColor,anchor=base,inner sep=0pt, outer sep=0pt, scale=  0.80] at (169.74, 12.20) {XLM-R};

\node[text=drawColor,anchor=base,inner sep=0pt, outer sep=0pt, scale=  0.80] at (169.74,  3.56) {Large};

\node[text=drawColor,anchor=base,inner sep=0pt, outer sep=0pt, scale=  0.80] at (255.80, 12.20) {Legal-XLM-R};

\node[text=drawColor,anchor=base,inner sep=0pt, outer sep=0pt, scale=  0.80] at (255.80,  3.56) {Base};

\node[text=drawColor,anchor=base,inner sep=0pt, outer sep=0pt, scale=  0.80] at (341.85, 12.20) {Legal-XLM-R};

\node[text=drawColor,anchor=base,inner sep=0pt, outer sep=0pt, scale=  0.80] at (341.85,  3.56) {Large};
\end{scope}
\begin{scope}
\path[clip] (  0.00,  0.00) rectangle (397.48,216.81);
\definecolor{drawColor}{RGB}{0,0,0}

\node[text=drawColor,rotate= 90.00,anchor=base,inner sep=0pt, outer sep=0pt, scale=  1.00] at (  8.89,118.51) {Degradation (pp)};
\end{scope}
\begin{scope}
\path[clip] (  0.00,  0.00) rectangle (397.48,216.81);
\definecolor{fillColor}{RGB}{255,255,255}

\path[fill=fillColor] (296.77,172.24) rectangle (381.77,211.15);
\end{scope}
\begin{scope}
\path[clip] (  0.00,  0.00) rectangle (397.48,216.81);
\definecolor{fillColor}{RGB}{215,48,39}

\path[fill=fillColor] (302.42,192.34) rectangle (315.58,205.50);
\end{scope}
\begin{scope}
\path[clip] (  0.00,  0.00) rectangle (397.48,216.81);
\definecolor{fillColor}{RGB}{69,117,180}

\path[fill=fillColor] (302.42,177.89) rectangle (315.58,191.05);
\end{scope}
\begin{scope}
\path[clip] (  0.00,  0.00) rectangle (397.48,216.81);
\definecolor{drawColor}{RGB}{0,0,0}

\node[text=drawColor,anchor=base west,inner sep=0pt, outer sep=0pt, scale=  0.80] at (321.23,196.17) {Forward $\Delta_{\mathrm{fwd}}$};
\end{scope}
\begin{scope}
\path[clip] (  0.00,  0.00) rectangle (397.48,216.81);
\definecolor{drawColor}{RGB}{0,0,0}

\node[text=drawColor,anchor=base west,inner sep=0pt, outer sep=0pt, scale=  0.80] at (321.23,181.72) {Backward $\Delta_{\mathrm{bwd}}$};
\end{scope}
\end{tikzpicture}
\caption{Forward and backward degradation by model. General XLM-R models show severe forward degradation but \emph{negative} backward degradation (improvement). Legal models degrade moderately in both directions. The dashed line separates general from legal-domain models.}
\label{fig:asymmetry}
\end{figure}

The $2 \times 2$ design allows us to isolate the effect of legal-domain pretraining from model scale.
Table~\ref{tab:degradation} and Figure~\ref{fig:asymmetry} summarize forward degradation, backward degradation, and asymmetry for all four models.

\begin{table}[t]
\centering
\caption{Forward--backward degradation and asymmetry. Forward: $F1(E_1{\to}E_1) - F1(E_1{\to}E_3)$. Backward: $F1(E_3{\to}E_3) - F1(E_3{\to}E_1)$. Positive backward values indicate degradation; negative indicates improvement.}
\label{tab:degradation}
\begin{tabular}{lccc}
\toprule
\textbf{Model} & \textbf{Fwd $\Delta$} & \textbf{Bwd $\Delta$} & \textbf{Asymmetry} \\
\midrule
XLM-R Base          & +20.3 & $-$9.0  & 29.3 \\
XLM-R Large         & +27.2 & $-$9.6  & 36.8 \\
Legal-XLM-R Base    & +11.2 & +3.8  & 7.4 \\
Legal-XLM-R Large   & +12.2 & +5.5  & 6.7 \\
\bottomrule
\end{tabular}
\end{table}

Legal-domain pretraining has a paradoxical effect on temporal robustness.
On one hand, Legal-XLM-R models show \emph{substantially less forward degradation} than their general counterparts: 11--12 pp vs.\ 20--27 pp.
On the other hand, this apparent robustness is achieved at the cost of much lower absolute performance (55.9/57.4 vs.\ 63.4/64.7 mean in-epoch F1).
Legal-XLM-R models degrade less because they start from a lower baseline -- their representations are less epoch-specific but also less discriminative.

The backward transfer pattern reveals a deeper distinction.
General XLM-R models \emph{improve} when transferring backward (negative $\Delta_\text{bwd}$): the full-scale model does better on pre-war data than on its own epoch.
Legal models, by contrast, \emph{degrade} in both directions ($\Delta_\text{bwd}$ of +3.8 and +5.5), suggesting that legal pretraining produces representations that are more brittle to any temporal shift, whether forward or backward.

The asymmetry gap provides the clearest contrast: 29.3--36.8 for general models vs.\ 6.7--7.4 for legal models.
General models exhibit strongly directional temporal transfer; legal models degrade more symmetrically but from a lower baseline.

\subsection{Forward--Backward Asymmetry}

The asymmetric degradation pattern -- severe forward, mild or negative backward -- is consistent with the ``legal language is additive'' hypothesis from Ovcharov~\cite{ovcharov2025tokenizer}.
Legislative change introduces new statutes, procedural rules, and legal concepts, but rarely abolishes existing ones entirely.
A model trained on pre-war data has never seen martial law provisions, collaborationism charges, or wartime procedural timelines; it lacks the representations needed to classify full-scale era decisions.
Conversely, a model trained on full-scale data has seen decisions that reference both old and new legislation, giving it adequate representations for pre-war classification.

\begin{figure}[t]
\centering
\begin{tikzpicture}[x=1pt,y=1pt]
\definecolor{fillColor}{RGB}{255,255,255}
\path[use as bounding box,fill=fillColor,fill opacity=0.00] (0,0) rectangle (397.48,216.81);
\begin{scope}
\path[clip] (  0.00,  0.00) rectangle (397.48,216.81);
\definecolor{fillColor}{RGB}{255,255,255}

\path[fill=fillColor] (  0.00,  0.00) rectangle (397.48,216.81);
\end{scope}
\begin{scope}
\path[clip] ( 25.83, 49.27) rectangle (207.16,198.85);
\definecolor{drawColor}{gray}{0.92}

\path[draw=drawColor,line width= 0.5pt,line join=round] ( 25.83, 56.07) --
	(207.16, 56.07);

\path[draw=drawColor,line width= 0.5pt,line join=round] ( 25.83, 90.06) --
	(207.16, 90.06);

\path[draw=drawColor,line width= 0.5pt,line join=round] ( 25.83,124.06) --
	(207.16,124.06);

\path[draw=drawColor,line width= 0.5pt,line join=round] ( 25.83,158.06) --
	(207.16,158.06);

\path[draw=drawColor,line width= 0.5pt,line join=round] ( 25.83,192.05) --
	(207.16,192.05);
\definecolor{fillColor}{RGB}{33,102,172}

\path[fill=fillColor] ( 41.41, 56.07) rectangle ( 58.41,171.48);
\definecolor{fillColor}{RGB}{215,48,39}

\path[fill=fillColor] ( 61.24, 56.07) rectangle ( 78.24,137.15);
\definecolor{fillColor}{RGB}{33,102,172}

\path[fill=fillColor] ( 98.08, 56.07) rectangle (115.08,169.95);
\definecolor{fillColor}{RGB}{215,48,39}

\path[fill=fillColor] (117.91, 56.07) rectangle (134.91,132.90);
\definecolor{fillColor}{RGB}{33,102,172}

\path[fill=fillColor] (154.74, 56.07) rectangle (171.74,155.00);
\definecolor{fillColor}{RGB}{215,48,39}

\path[fill=fillColor] (174.57, 56.07) rectangle (191.57,137.32);
\end{scope}
\begin{scope}
\path[clip] (212.16, 49.27) rectangle (393.48,198.85);
\definecolor{drawColor}{gray}{0.92}

\path[draw=drawColor,line width= 0.5pt,line join=round] (212.16, 56.07) --
	(393.48, 56.07);

\path[draw=drawColor,line width= 0.5pt,line join=round] (212.16, 90.06) --
	(393.48, 90.06);

\path[draw=drawColor,line width= 0.5pt,line join=round] (212.16,124.06) --
	(393.48,124.06);

\path[draw=drawColor,line width= 0.5pt,line join=round] (212.16,158.06) --
	(393.48,158.06);

\path[draw=drawColor,line width= 0.5pt,line join=round] (212.16,192.05) --
	(393.48,192.05);
\definecolor{fillColor}{RGB}{33,102,172}

\path[fill=fillColor] (227.74, 56.07) rectangle (244.74,177.94);
\definecolor{fillColor}{RGB}{215,48,39}

\path[fill=fillColor] (247.57, 56.07) rectangle (264.57,136.13);
\definecolor{fillColor}{RGB}{33,102,172}

\path[fill=fillColor] (284.40, 56.07) rectangle (301.40,165.53);
\definecolor{fillColor}{RGB}{215,48,39}

\path[fill=fillColor] (304.24, 56.07) rectangle (321.24,122.19);
\definecolor{fillColor}{RGB}{33,102,172}

\path[fill=fillColor] (341.07, 56.07) rectangle (358.07,166.72);
\definecolor{fillColor}{RGB}{215,48,39}

\path[fill=fillColor] (360.90, 56.07) rectangle (377.90,119.13);
\end{scope}
\begin{scope}
\path[clip] ( 25.83,198.85) rectangle (207.16,214.81);
\definecolor{drawColor}{gray}{0.10}

\node[text=drawColor,anchor=base,inner sep=0pt, outer sep=0pt, scale=  0.90] at (116.49,203.72) {\bfseries XLM-R Base};
\end{scope}
\begin{scope}
\path[clip] (212.16,198.85) rectangle (393.48,214.81);
\definecolor{drawColor}{gray}{0.10}

\node[text=drawColor,anchor=base,inner sep=0pt, outer sep=0pt, scale=  0.90] at (302.82,203.72) {\bfseries XLM-R Large};
\end{scope}
\begin{scope}
\path[clip] (  0.00,  0.00) rectangle (397.48,216.81);
\definecolor{drawColor}{gray}{0.30}

\node[text=drawColor,anchor=base,inner sep=0pt, outer sep=0pt, scale=  0.80] at ( 59.83, 39.26) {Approved};

\node[text=drawColor,anchor=base,inner sep=0pt, outer sep=0pt, scale=  0.80] at (116.49, 39.26) {Dismissed};

\node[text=drawColor,anchor=base,inner sep=0pt, outer sep=0pt, scale=  0.80] at (173.16, 39.26) {Partial};
\end{scope}
\begin{scope}
\path[clip] (  0.00,  0.00) rectangle (397.48,216.81);
\definecolor{drawColor}{gray}{0.30}

\node[text=drawColor,anchor=base,inner sep=0pt, outer sep=0pt, scale=  0.80] at (246.16, 39.26) {Approved};

\node[text=drawColor,anchor=base,inner sep=0pt, outer sep=0pt, scale=  0.80] at (302.82, 39.26) {Dismissed};

\node[text=drawColor,anchor=base,inner sep=0pt, outer sep=0pt, scale=  0.80] at (359.49, 39.26) {Partial};
\end{scope}
\begin{scope}
\path[clip] (  0.00,  0.00) rectangle (397.48,216.81);
\definecolor{drawColor}{gray}{0.30}

\node[text=drawColor,anchor=base east,inner sep=0pt, outer sep=0pt, scale=  0.80] at ( 21.33, 53.31) {0};

\node[text=drawColor,anchor=base east,inner sep=0pt, outer sep=0pt, scale=  0.80] at ( 21.33, 87.31) {20};

\node[text=drawColor,anchor=base east,inner sep=0pt, outer sep=0pt, scale=  0.80] at ( 21.33,121.30) {40};

\node[text=drawColor,anchor=base east,inner sep=0pt, outer sep=0pt, scale=  0.80] at ( 21.33,155.30) {60};

\node[text=drawColor,anchor=base east,inner sep=0pt, outer sep=0pt, scale=  0.80] at ( 21.33,189.30) {80};
\end{scope}
\begin{scope}
\path[clip] (  0.00,  0.00) rectangle (397.48,216.81);
\definecolor{drawColor}{RGB}{0,0,0}

\node[text=drawColor,rotate= 90.00,anchor=base,inner sep=0pt, outer sep=0pt, scale=  1.00] at (  8.89,124.06) {Per-class F1 (\%)};
\end{scope}
\begin{scope}
\path[clip] (  0.00,  0.00) rectangle (397.48,216.81);
\definecolor{fillColor}{RGB}{33,102,172}

\path[fill=fillColor] (133.43,  7.65) rectangle (146.59, 22.06);
\end{scope}
\begin{scope}
\path[clip] (  0.00,  0.00) rectangle (397.48,216.81);
\definecolor{fillColor}{RGB}{215,48,39}

\path[fill=fillColor] (188.10,  7.65) rectangle (201.26, 22.06);
\end{scope}
\begin{scope}
\path[clip] (  0.00,  0.00) rectangle (397.48,216.81);
\definecolor{drawColor}{RGB}{0,0,0}

\node[text=drawColor,anchor=base west,inner sep=0pt, outer sep=0pt, scale=  0.80] at (152.24, 12.10) {In-epoch};
\end{scope}
\begin{scope}
\path[clip] (  0.00,  0.00) rectangle (397.48,216.81);
\definecolor{drawColor}{RGB}{0,0,0}

\node[text=drawColor,anchor=base west,inner sep=0pt, outer sep=0pt, scale=  0.80] at (206.90, 16.42) {Cross-epoch};

\node[text=drawColor,anchor=base west,inner sep=0pt, outer sep=0pt, scale=  0.80] at (206.90,  7.78) {(Pre-war $\to$ Full-scale)};
\end{scope}
\end{tikzpicture}
\caption{Per-class F1 for in-epoch vs.\ cross-epoch (pre-war $\to$ full-scale) evaluation. \textit{Dismissed} suffers the largest drop for both models. XLM-R Large shows uniformly steeper degradation, consistent with stronger epoch overfitting.}
\label{fig:per_class}
\end{figure}
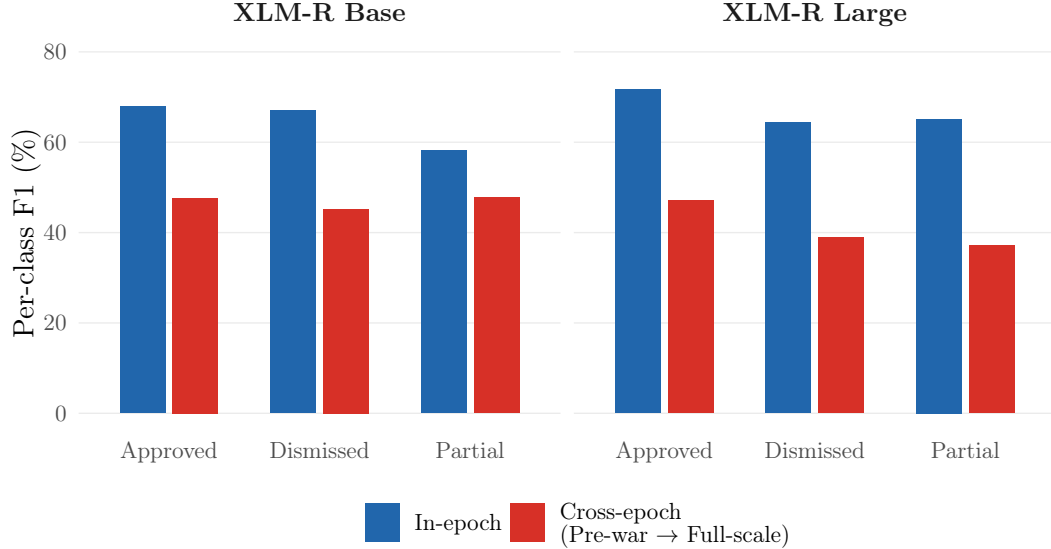

The per-class analysis supports this interpretation.
For XLM-R Base trained on pre-war and tested on full-scale, the \textit{dismissed} class suffers the steepest F1 drop ($-$21.8 pp), while \textit{partial} is the most resilient ($-$10.4 pp; Figure~\ref{fig:per_class}).
Dismissal reasoning underwent the greatest structural change between epochs -- new grounds for dismissal under martial law are categorically absent from pre-war training data.

The scale effect amplifies asymmetry.
XLM-R Large exhibits 36.8 asymmetry gap vs.\ 29.3 for Base, consistent with larger models overfitting more to distributional features of their training epoch.
Legal pretraining compresses the asymmetry gap to 6.7--7.4, but as shown above, this comes at the cost of absolute performance.

\subsection{Continual Learning}

The cross-epoch generalization results (Section~5.2) establish that temporal drift degrades performance.
A natural follow-up question is whether sequential fine-tuning across epochs -- continual learning -- can mitigate this degradation without catastrophic forgetting~\cite{kirkpatrick2017overcoming}.
We evaluate both chronological (forward: pre-war $\to$ hybrid $\to$ full-scale) and reverse-chronological (backward: full-scale $\to$ hybrid $\to$ pre-war) training orders.

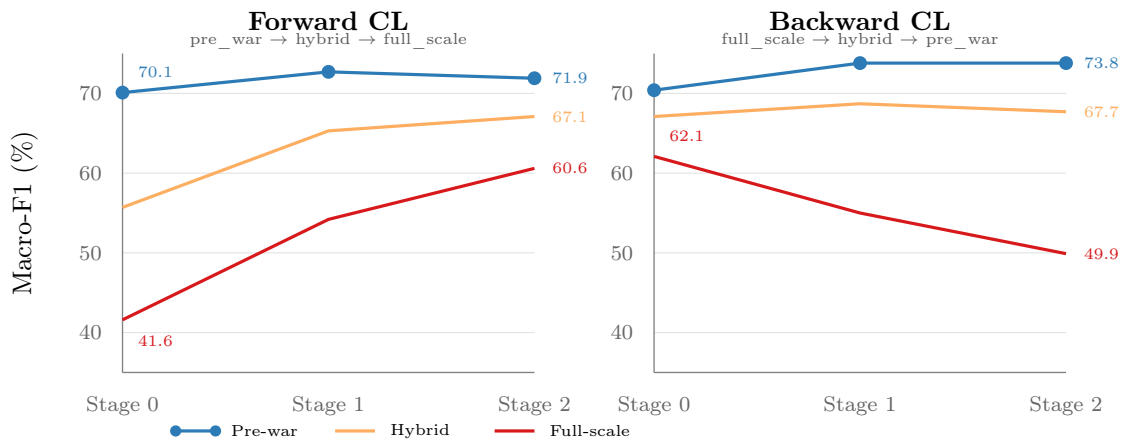
\begin{figure}[t]
\centering
\begin{tikzpicture}[x=1pt,y=1pt]
\definecolor{colPre}{RGB}{44,123,182}
\definecolor{colHyb}{RGB}{253,174,97}
\definecolor{colFs}{RGB}{215,25,28}
\definecolor{gridColor}{gray}{0.88}

\def\pw{155}   
\def\ph{120}   
\def\gap{45}   
\def\xoff{40}  

\def\yscale{3}
\def\ymin{35}

\def\xs{0}
\def\xm{77.5}
\def\xe{155}

\begin{scope}[shift={(\xoff,0)}]
  \foreach \y in {40,50,60,70} {
    \pgfmathsetmacro{\yp}{(\y-\ymin)*\yscale}
    \draw[gridColor, line width=0.3pt] (0,\yp) -- (\pw,\yp);
  }
  \draw[gray, line width=0.5pt] (0,0) -- (\pw,0);
  \draw[gray, line width=0.5pt] (0,0) -- (0,\ph);
  \foreach \y in {40,50,60,70} {
    \pgfmathsetmacro{\yp}{(\y-\ymin)*\yscale}
    \node[anchor=east, font=\scriptsize, gray!80!black] at (-4,\yp) {\y};
  }
  \node[anchor=north, font=\scriptsize, gray!80!black] at (\xs,-5) {Stage 0};
  \node[anchor=north, font=\scriptsize, gray!80!black] at (\xm,-5) {Stage 1};
  \node[anchor=north, font=\scriptsize, gray!80!black] at (\xe,-5) {Stage 2};

  \draw[colPre, line width=1.2pt, mark=*, mark size=2pt]
    plot coordinates {
      (\xs,{(70.1-\ymin)*\yscale})
      (\xm,{(72.7-\ymin)*\yscale})
      (\xe,{(71.9-\ymin)*\yscale})
    };
  \draw[colHyb, line width=1.2pt, mark=square*, mark size=2pt]
    plot coordinates {
      (\xs,{(55.7-\ymin)*\yscale})
      (\xm,{(65.3-\ymin)*\yscale})
      (\xe,{(67.1-\ymin)*\yscale})
    };
  \draw[colFs, line width=1.2pt, mark=triangle*, mark size=2.5pt]
    plot coordinates {
      (\xs,{(41.6-\ymin)*\yscale})
      (\xm,{(54.2-\ymin)*\yscale})
      (\xe,{(60.6-\ymin)*\yscale})
    };

  \node[colPre, font=\tiny, anchor=west] at (\xe+3,{(71.9-\ymin)*\yscale}) {71.9};
  \node[colHyb, font=\tiny, anchor=west] at (\xe+3,{(67.1-\ymin)*\yscale}) {67.1};
  \node[colFs, font=\tiny, anchor=west] at (\xe+3,{(60.6-\ymin)*\yscale}) {60.6};

  \node[colPre, font=\tiny, anchor=south west] at (\xs+2,{(70.1-\ymin)*\yscale+2}) {70.1};
  \node[colFs, font=\tiny, anchor=north west] at (\xs+2,{(41.6-\ymin)*\yscale-2}) {41.6};

  \node[anchor=south, font=\small\bfseries] at (\pw/2, \ph+5) {Forward CL};
  \node[anchor=south, font=\tiny, gray!70!black] at (\pw/2, \ph-2) {pre\_war $\to$ hybrid $\to$ full\_scale};
\end{scope}

\begin{scope}[shift={(\xoff+\pw+\gap,0)}]
  \foreach \y in {40,50,60,70} {
    \pgfmathsetmacro{\yp}{(\y-\ymin)*\yscale}
    \draw[gridColor, line width=0.3pt] (0,\yp) -- (\pw,\yp);
  }
  \draw[gray, line width=0.5pt] (0,0) -- (\pw,0);
  \draw[gray, line width=0.5pt] (0,0) -- (0,\ph);
  \foreach \y in {40,50,60,70} {
    \pgfmathsetmacro{\yp}{(\y-\ymin)*\yscale}
    \node[anchor=east, font=\scriptsize, gray!80!black] at (-4,\yp) {\y};
  }
  \node[anchor=north, font=\scriptsize, gray!80!black] at (\xs,-5) {Stage 0};
  \node[anchor=north, font=\scriptsize, gray!80!black] at (\xm,-5) {Stage 1};
  \node[anchor=north, font=\scriptsize, gray!80!black] at (\xe,-5) {Stage 2};

  \draw[colPre, line width=1.2pt, mark=*, mark size=2pt]
    plot coordinates {
      (\xs,{(70.4-\ymin)*\yscale})
      (\xm,{(73.8-\ymin)*\yscale})
      (\xe,{(73.8-\ymin)*\yscale})
    };
  \draw[colHyb, line width=1.2pt, mark=square*, mark size=2pt]
    plot coordinates {
      (\xs,{(67.1-\ymin)*\yscale})
      (\xm,{(68.7-\ymin)*\yscale})
      (\xe,{(67.7-\ymin)*\yscale})
    };
  \draw[colFs, line width=1.2pt, mark=triangle*, mark size=2.5pt]
    plot coordinates {
      (\xs,{(62.1-\ymin)*\yscale})
      (\xm,{(55.0-\ymin)*\yscale})
      (\xe,{(49.9-\ymin)*\yscale})
    };

  \node[colPre, font=\tiny, anchor=west] at (\xe+3,{(73.8-\ymin)*\yscale}) {73.8};
  \node[colHyb, font=\tiny, anchor=west] at (\xe+3,{(67.7-\ymin)*\yscale}) {67.7};
  \node[colFs, font=\tiny, anchor=west] at (\xe+3,{(49.9-\ymin)*\yscale}) {49.9};

  \node[colFs, font=\tiny, anchor=south west] at (\xs+2,{(62.1-\ymin)*\yscale+2}) {62.1};

  \node[anchor=south, font=\small\bfseries] at (\pw/2, \ph+5) {Backward CL};
  \node[anchor=south, font=\tiny, gray!70!black] at (\pw/2, \ph-2) {full\_scale $\to$ hybrid $\to$ pre\_war};
\end{scope}

\begin{scope}[shift={(\xoff + \pw/2 + \gap/2, -22)}]
  \draw[colPre, line width=1.2pt, mark=*, mark size=1.5pt] plot coordinates {(-80,0) (-65,0)};
  \node[font=\tiny, anchor=west] at (-63,0) {Pre-war};
  \draw[colHyb, line width=1.2pt, mark=square*, mark size=1.5pt] plot coordinates {(-20,0) (-5,0)};
  \node[font=\tiny, anchor=west] at (-3,0) {Hybrid};
  \draw[colFs, line width=1.2pt, mark=triangle*, mark size=2pt] plot coordinates {(40,0) (55,0)};
  \node[font=\tiny, anchor=west] at (57,0) {Full-scale};
\end{scope}

\node[rotate=90, anchor=south, font=\small] at (12, 60) {Macro-F1 (\%)};

\end{tikzpicture}
\caption{Continual learning trajectories for XLM-R Large. Left: forward (chronological) training retains pre-war knowledge while steadily gaining full-scale performance. Right: backward (reverse-chronological) training causes catastrophic forgetting of full-scale knowledge ($-$12.2~pp) while pre-war remains stable. The directional asymmetry mirrors the cross-epoch generalization asymmetry from a complementary angle.}
\label{fig:cl_trajectory}
\end{figure}

Table~\ref{tab:cl_summary} and Figure~\ref{fig:cl_trajectory} present the results.

\begin{table}[t]
\centering
\caption{Continual learning: origin retention and target acquisition. Origin = first-trained epoch; Target = last-trained epoch. Forward: origin is pre-war, target is full-scale. Backward: origin is full-scale, target is pre-war. $n$: number of seeds with full per-epoch metrics. Positive $\Delta$ = retention/gain; negative = forgetting.}
\label{tab:cl_summary}
\begin{tabular}{llc rr rr}
\toprule
\textbf{Model} & \textbf{Dir.} & $n$ & \multicolumn{2}{c}{\textbf{Origin F1}} & \multicolumn{2}{c}{\textbf{Target F1}} \\
\cmidrule(lr){4-5} \cmidrule(lr){6-7}
 & & & Start & $\Delta$ & Start & $\Delta$ \\
\midrule
\multirow{2}{*}{XLM-R Base}
  & Fwd & 2 & 65.0 & $+$6.2  & 43.2 & $+$16.5 \\
  & Bwd & 1 & 59.5 & $-$14.3 & 69.0 & $-$0.6 \\
\midrule
\multirow{2}{*}{XLM-R Large}
  & Fwd & 2 & 70.1 & $+$1.8  & 41.6 & $+$19.0 \\
  & Bwd & 1 & 62.1 & $-$12.2 & 70.4 & $+$3.4 \\
\midrule
Legal-XLM-R Large
  & Fwd & 2 & 49.2 & $-$5.6  & 40.4 & $+$16.8 \\
\bottomrule
\end{tabular}
\end{table}

\paragraph{Forward continual learning eliminates catastrophic forgetting for general models.}
When XLM-R models are fine-tuned chronologically (pre-war $\to$ hybrid $\to$ full-scale), origin-epoch knowledge is not only preserved but sometimes \emph{improved}: XLM-R Base gains +6.2~pp on pre-war data after all three stages, while XLM-R Large retains within +1.8~pp.
Simultaneously, full-scale performance increases dramatically: +16.5~pp for Base and +19.0~pp for Large.
The resulting model after forward continual learning achieves a mean cross-epoch F1 of 66.5 (XLM-R Large) -- surpassing the best single-epoch model (hybrid-trained XLM-R Large at 63.7 mean) by 2.8~pp, with particularly large gains on full-scale (+9.2~pp over the best single-epoch off-diagonal score of 51.4).

\paragraph{Backward continual learning causes catastrophic forgetting.}
Reverse-chronological training (full-scale $\to$ hybrid $\to$ pre-war) produces the opposite pattern.
XLM-R Base loses 14.3~pp on full-scale data -- more than the entire forward degradation gap from Section~5.2.
XLM-R Large loses 12.2~pp, with full-scale F1 dropping from 62.1 to 49.9 over three stages (Figure~\ref{fig:cl_trajectory}, right panel).
Pre-war knowledge, by contrast, remains stable or improves slightly (+3.4~pp for Large), mirroring the backward transfer robustness observed in single-epoch experiments.

\paragraph{Legal-domain pretraining is incompatible with continual learning.}
Legal-XLM-R Large loses 5.6~pp on pre-war data even under \emph{forward} continual learning -- the direction where general models retain perfectly.
This extends the finding from Section~5.3: legal pretraining produces representations that are not only less discriminative but also more fragile to sequential distribution shifts.
Full per-epoch backward CL data for Legal-XLM-R was not available; however, the forward fragility alone disqualifies legal-domain models from continual learning pipelines on jurisdiction-specific tasks.

\paragraph{The asymmetry reverses for continual learning.}
In cross-epoch generalization (Section~5.4), backward transfer was the robust direction: models trained on full-scale data generalized well to pre-war test data.
In continual learning, the robust direction is \emph{forward}: chronological training preserves prior knowledge, while reverse-chronological training destroys it.
Both asymmetries are explained by the same mechanism -- legal language is additive -- but the mechanism operates differently.
In single-epoch training, a model trained on newer (richer) data already contains representations for older (simpler) distributions.
In sequential training, chronological order presents progressively more complex distributions, allowing the model to \emph{extend} its representations rather than \emph{overwrite} them; reverse order forces the model to learn simpler distributions that do not reinforce the complex representations acquired earlier.

\subsection{Cross-Jurisdictional Temporal Transfer}

The experiments above establish temporal drift within a single jurisdiction.
A natural question is whether cross-jurisdictional pretraining can mitigate temporal degradation by providing more diverse legal representations.
We test this using Swiss Judgment Prediction (SJP)~\cite{niklaus2021swiss}, extending the Cross-X transfer framework~\cite{niklaus2022cross} to the temporal dimension.

\paragraph{Zero-shot transfer is near-random.}
An XLM-R Base model trained on SJP (all languages, binary) and tested on Ukrainian (binary, dropping partial) achieves 33.8--35.7\% macro-F1 across epochs -- below the 50\% binary chance level.
The temporal ordering is weakly preserved (hybrid 35.7 $>$ pre-war 33.8 $>$ full-scale 32.6), confirming that full-scale data is the hardest target even for a model with no Ukrainian training data.
The SJP model achieves 49.4\% on its own test set, indicating that cross-jurisdictional transfer is nearly non-existent at the zero-shot level.

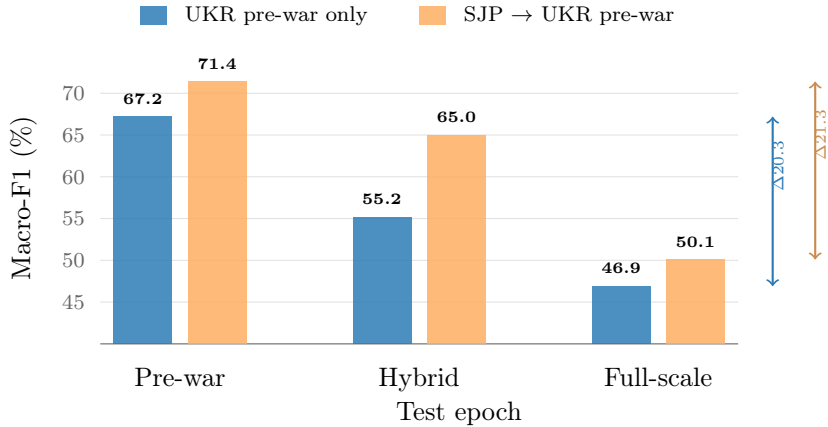
\begin{figure}[t]
\centering
\begin{tikzpicture}[x=1pt,y=1pt]
\definecolor{colUKR}{RGB}{44,123,182}
\definecolor{colSJP}{RGB}{253,174,97}
\definecolor{gridColor}{gray}{0.88}

\def\barw{22}     
\def\barsep{6}    
\def\groupsep{40} 
\def\ph{110}      
\def\xoff{35}     

\pgfmathsetmacro{\yscale}{\ph/35}
\def\ymin{40}

\pgfmathsetmacro{\gA}{0}
\pgfmathsetmacro{\gB}{\gA + 2*\barw + \barsep + \groupsep}
\pgfmathsetmacro{\gC}{\gB + 2*\barw + \barsep + \groupsep}

\begin{scope}[shift={(\xoff,0)}]
  \foreach \y in {45,50,55,60,65,70} {
    \pgfmathsetmacro{\yp}{(\y-\ymin)*\yscale}
    \draw[gridColor, line width=0.3pt] (-5,\yp) -- ({\gC+2*\barw+\barsep+5},\yp);
  }
  \draw[gray, line width=0.5pt] (-5,0) -- ({\gC+2*\barw+\barsep+5},0);

  \foreach \y in {45,50,55,60,65,70} {
    \pgfmathsetmacro{\yp}{(\y-\ymin)*\yscale}
    \node[anchor=east, font=\scriptsize, gray!80!black] at (-7,\yp) {\y};
  }

  \pgfmathsetmacro{\hA}{(67.2-\ymin)*\yscale}
  \fill[colUKR, opacity=0.85] (\gA,0) rectangle ({\gA+\barw},\hA);
  \node[font=\tiny\bfseries, anchor=south] at ({\gA+\barw/2},\hA+1) {67.2};
  \pgfmathsetmacro{\hB}{(71.4-\ymin)*\yscale}
  \fill[colSJP, opacity=0.85] ({\gA+\barw+\barsep},0) rectangle ({\gA+2*\barw+\barsep},\hB);
  \node[font=\tiny\bfseries, anchor=south] at ({\gA+\barw+\barsep+\barw/2},\hB+1) {71.4};
  \node[font=\small, anchor=north] at ({\gA+\barw+\barsep/2},-5) {Pre-war};

  \pgfmathsetmacro{\hA}{(55.2-\ymin)*\yscale}
  \fill[colUKR, opacity=0.85] (\gB,0) rectangle ({\gB+\barw},\hA);
  \node[font=\tiny\bfseries, anchor=south] at ({\gB+\barw/2},\hA+1) {55.2};
  \pgfmathsetmacro{\hB}{(65.0-\ymin)*\yscale}
  \fill[colSJP, opacity=0.85] ({\gB+\barw+\barsep},0) rectangle ({\gB+2*\barw+\barsep},\hB);
  \node[font=\tiny\bfseries, anchor=south] at ({\gB+\barw+\barsep+\barw/2},\hB+1) {65.0};
  \node[font=\small, anchor=north] at ({\gB+\barw+\barsep/2},-5) {Hybrid};

  \pgfmathsetmacro{\hA}{(46.9-\ymin)*\yscale}
  \fill[colUKR, opacity=0.85] (\gC,0) rectangle ({\gC+\barw},\hA);
  \node[font=\tiny\bfseries, anchor=south] at ({\gC+\barw/2},\hA+1) {46.9};
  \pgfmathsetmacro{\hB}{(50.1-\ymin)*\yscale}
  \fill[colSJP, opacity=0.85] ({\gC+\barw+\barsep},0) rectangle ({\gC+2*\barw+\barsep},\hB);
  \node[font=\tiny\bfseries, anchor=south] at ({\gC+\barw+\barsep+\barw/2},\hB+1) {50.1};
  \node[font=\small, anchor=north] at ({\gC+\barw+\barsep/2},-5) {Full-scale};

  \pgfmathsetmacro{\yUKRpre}{(67.2-\ymin)*\yscale}
  \pgfmathsetmacro{\yUKRfs}{(46.9-\ymin)*\yscale}
  \pgfmathsetmacro{\ySJPpre}{(71.4-\ymin)*\yscale}
  \pgfmathsetmacro{\ySJPfs}{(50.1-\ymin)*\yscale}

  \pgfmathsetmacro{\xAnnot}{\gC+2*\barw+\barsep+18}
  \draw[colUKR, line width=0.8pt, <->] (\xAnnot,\yUKRpre) -- (\xAnnot,\yUKRfs);
  \node[colUKR, font=\tiny, anchor=west, rotate=90] at (\xAnnot+2,{(\yUKRpre+\yUKRfs)/2}) {$\Delta$20.3};

  \pgfmathsetmacro{\xAnnotB}{\xAnnot+16}
  \draw[colSJP!80!black, line width=0.8pt, <->] (\xAnnotB,\ySJPpre) -- (\xAnnotB,\ySJPfs);
  \node[colSJP!80!black, font=\tiny, anchor=west, rotate=90] at (\xAnnotB+2,{(\ySJPpre+\ySJPfs)/2}) {$\Delta$21.3};

\end{scope}

\node[rotate=90, anchor=south, font=\small] at (10, 55) {Macro-F1 (\%)};

\node[font=\small, anchor=north] at (165,-18) {Test epoch};

\fill[colUKR, opacity=0.85] (45,120) rectangle (55,127);
\node[font=\scriptsize, anchor=west] at (58,123.5) {UKR pre-war only};
\fill[colSJP, opacity=0.85] (150,120) rectangle (160,127);
\node[font=\scriptsize, anchor=west] at (163,123.5) {SJP $\to$ UKR pre-war};

\end{tikzpicture}
\caption{Cross-jurisdictional pretraining effect. SJP pretraining lifts all epochs (+3.2 to +9.8~pp) but the forward degradation gap is unchanged ($\Delta$20.3 vs.\ $\Delta$21.3~pp). Temporal drift is orthogonal to cross-jurisdictional transfer.}
\label{fig:cross_jurisdiction}
\end{figure}

\paragraph{Cross-jurisdictional pretraining lifts absolute performance but preserves temporal degradation.}
Table~\ref{tab:cross_jurisdiction} and Figure~\ref{fig:cross_jurisdiction} present the key comparison.
When XLM-R Base is first fine-tuned on SJP (binary) and then on Ukrainian pre-war (3-class), performance improves at every epoch compared to Ukrainian-only training: +4.2~pp on pre-war, +9.8~pp on hybrid, and +3.2~pp on full-scale.
However, the forward degradation magnitude is virtually unchanged: 21.3~pp (transfer) vs.\ 20.3~pp (Ukrainian-only).
Cross-jurisdictional pretraining shifts the entire performance curve upward without altering its slope.

\begin{table}[t]
\centering
\caption{Effect of cross-jurisdictional pretraining on temporal robustness (XLM-R Base, pre-war fine-tuned, 3-class). SJP$\to$UKR: Phase~1 on Swiss JP, Phase~2 on Ukrainian pre-war. Mean $\pm$ std over 3 seeds.}
\label{tab:cross_jurisdiction}
\begin{tabular}{l ccc c}
\toprule
\textbf{Training} & \textbf{Pre-war} & \textbf{Hybrid} & \textbf{Full-scale} & \textbf{Fwd $\Delta$} \\
\midrule
UKR only        & 67.2{\small$\pm$0.8} & 55.2{\small$\pm$1.1} & 46.9{\small$\pm$4.1} & 20.3 \\
SJP$\to$UKR     & 71.4{\small$\pm$1.0} & 65.0{\small$\pm$0.5} & 50.1{\small$\pm$2.2} & 21.3 \\
\midrule
Improvement     & $+$4.2 & $+$9.8 & $+$3.2 & $-$1.0 \\
\bottomrule
\end{tabular}
\end{table}

\paragraph{Reverse transfer: Ukrainian to Swiss.}
Training on Ukrainian hybrid-war (binary) and testing on SJP yields 45.4\% macro-F1 on SJP (vs.\ 77.9\% on Ukrainian self-test), with consistent results across SJP languages (German 45.0\%, French 45.9\%, Italian 45.0\%).
The jurisdiction gap is roughly symmetric: SJP$\to$UKR gives $\sim$34\% on Ukrainian; UKR$\to$SJP gives $\sim$45\% on Swiss.

\paragraph{Temporal drift is jurisdiction-independent.}
These results demonstrate that temporal drift is not a jurisdiction-specific artifact.
The same degradation magnitude persists regardless of whether the model receives cross-jurisdictional pretraining, confirming that temporal concept drift in legal NLP arises from the evolution of legal language itself -- legislative change, procedural reform, and doctrinal development -- rather than from limitations of the training data.
This finding extends the Cross-X framework~\cite{niklaus2022cross}: whereas Niklaus et al.\ showed that cross-lingual, cross-domain, and cross-regional transfer can partially bridge jurisdiction gaps, we show that cross-temporal degradation is resistant to such transfer.

\section{Discussion}

\paragraph{Temporal drift exceeds all other factors.}
The forward degradation gap of 20.3--27.2 pp (Table~\ref{tab:degradation}) exceeds the difference between the best and worst model on any in-epoch evaluation (12.8 pp), the gap between general and legal-domain pretraining (7--9 pp), and the improvement from cross-jurisdictional pretraining (+3--10 pp; Table~\ref{tab:cross_jurisdiction}).
This means that \emph{when} a model was trained matters more than \emph{which} model was chosen, \emph{how} it was pretrained, or \emph{what} foreign data it saw -- a finding with direct implications for deployment and maintenance of legal NLP systems.

\paragraph{Legal-domain pretraining and temporal robustness.}
Legal-XLM-R was pretrained on the MultiLegalPile, which spans multiple time periods and jurisdictions.
Despite this temporal diversity in pretraining data, legal models substantially underperform general XLM-R on Ukrainian court decisions.
We hypothesize two contributing factors: (1)~the MultiLegalPile is predominantly English-language, providing limited benefit for Ukrainian Cyrillic text, and (2)~legal pretraining may introduce domain-specific inductive biases that conflict with jurisdiction-specific patterns in Ukrainian civil law.
The reduced asymmetry gap (6.7--7.4 vs.\ 29.3--36.8) suggests that legal pretraining does produce more \emph{temporally uniform} representations, but at the cost of discriminative power.

\paragraph{Neural models vs.\ TF-IDF baselines.}
The substantial performance gap between neural models (63--65 mean F1) and TF-IDF baselines (79.8 mean F1) deserves explanation.
TF-IDF classifiers operate on the full text and capture keyword-level signals (e.g., explicit outcome phrases) that transformers cannot efficiently encode within a 512-token window on highly inflected, subword-fragmented Ukrainian text.
This result challenges the assumption that transformer encoders universally outperform classical methods -- for Ukrainian legal text, tokenizer fertility and sequence length limitations remain binding constraints.

\paragraph{Implications for benchmark design.}
Our results suggest that legal NLP benchmarks should report temporal robustness alongside aggregate performance.
A model that achieves 90\% macro-F1 on a randomly split benchmark may be substantially less reliable when deployed on decisions from a later period.
We propose that LEXTREME and future benchmarks include temporal split configurations alongside their standard evaluation.

\paragraph{Continual learning as mitigation.}
The continual learning results (Section~5.5) provide a direct mitigation strategy for temporal drift.
Forward (chronological) sequential fine-tuning yields models that outperform any single-epoch model in mean cross-epoch F1, gaining up to 19.0~pp on the most recent epoch while fully retaining knowledge of older epochs.
This suggests that legal NLP practitioners should maintain a single model and retrain it chronologically as new temporal data becomes available, rather than training separate models for each period.
The failure of backward CL provides a cautionary note: retraining order matters, and reverse-chronological curricula should be avoided.

\paragraph{Implications for practitioners.}
Given the severity of forward degradation (up to 27.2 pp), deployed legal NLP systems require periodic retraining on recent data.
Our continual learning results show that chronological retraining is not only safe (no catastrophic forgetting for general XLM-R) but beneficial: the resulting model surpasses any single-epoch model as a general-purpose classifier.
Backward-compatible models (trained on recent data, deployed on historical analysis) are substantially more reliable than forward-deployed models (trained on historical data, applied to new decisions).

\section{Limitations}

\paragraph{Cross-jurisdictional scope.}
While we include Swiss Judgment Prediction as a cross-jurisdictional source, the primary temporal evaluation is conducted on Ukrainian court decisions.
The three-epoch structure reflects Ukrainian-specific geopolitical disruptions; jurisdictions without comparable exogenous shocks may exhibit different degradation profiles.
A full temporal split of SJP (2000--2020) would enable direct comparison of degradation rates across jurisdictions.

\paragraph{Single task.}
We evaluate judgment prediction (3-class classification).
Other legal NLP tasks (NER, summarization, retrieval) may exhibit different temporal sensitivity profiles.

\paragraph{Encoder-only models.}
We focus on XLM-R variants, the standard LEXTREME evaluation architecture.
Decoder-based models (GPT-4, Claude, Llama) and encoder-decoder architectures may show different temporal robustness properties.

\paragraph{Epoch boundaries.}
While our epoch boundaries correspond to documented structural breaks, the choice of exact cutoff dates (2014-01-01, 2022-01-01) is somewhat arbitrary.
Sensitivity analysis with shifted boundaries would strengthen the findings.

\paragraph{Label noise.}
Outcome labels are extracted via regex from the dispositive section.
While validated on a 273-document subset~\cite{ovcharov2025tokenizer}, systematic errors in the regex parser could introduce epoch-dependent label noise.

\paragraph{Continual learning coverage.}
Our continual learning evaluation covers three models with 1--2 seeds per direction, fewer than the 3-seed protocol used for cross-epoch generalization.
Full per-epoch backward CL data for Legal-XLM-R Base and Legal-XLM-R Large backward is not available.
Additionally, we evaluate only naive sequential fine-tuning; regularization-based approaches (EWC~\cite{kirkpatrick2017overcoming}, experience replay) may further improve retention.

\paragraph{Embedding analysis.}
We do not evaluate embedding drift (CKA similarity~\cite{kornblith2019similarity} across epoch-trained models), which would complement our cross-epoch and continual learning findings by revealing whether temporal drift manifests as representational divergence or merely shifts in the classification head.

\section{Conclusion}

We have presented the first neural temporal robustness benchmark for legal NLP, fine-tuning four XLM-R variants on 428K Ukrainian court decisions across three temporal epochs.
Our cross-epoch generalization matrices reveal that temporal drift is a dominant source of performance degradation, exceeding the impact of model selection and domain pretraining.

Forward degradation is severe (up to 27.2 pp for XLM-R Large) and asymmetric.
The forward--backward asymmetry provides mechanistic insight: legal language is additive, with newer frameworks subsuming older ones but not vice versa.

Continual learning experiments reinforce this hypothesis from a complementary angle.
Chronological sequential fine-tuning (pre-war $\to$ hybrid $\to$ full-scale) eliminates catastrophic forgetting for general XLM-R models, retaining pre-war knowledge (+1.8 to +6.2~pp) while gaining up to +19.0~pp on full-scale data.
The resulting model surpasses any single-epoch model as a general-purpose classifier.
Reverse-chronological training causes severe forgetting ($-$12.2 to $-$14.3~pp), and Legal-XLM-R forgets even under chronological training, confirming that legal-domain pretraining produces temporally fragile representations for jurisdiction-specific tasks.

Cross-jurisdictional pretraining on Swiss Judgment Prediction data improves absolute performance by +3 to +10~pp but does not reduce temporal degradation magnitude, confirming that temporal drift is an intrinsic property of legal language evolution rather than a jurisdiction-specific artifact.

These findings have direct implications for practitioners: deployed legal NLP systems should be retrained chronologically as new temporal data becomes available, using general-purpose encoders rather than legal-domain pretrained variants.
Cross-jurisdictional pretraining can raise the performance baseline but cannot substitute for temporal retraining.

We release the full dataset (428K decisions, three epochs, chronological splits) as a LEXTREME contribution, providing the first Cyrillic-script legal NLP subset with temporal annotations.
We advocate for temporal robustness evaluation as a standard component of legal NLP benchmarking.

\paragraph{Data and code availability.}
The dataset is available at \url{https://huggingface.co/datasets/overthelex/ukrainian-court-decisions}.
Training scripts and evaluation code are released at \url{https://github.com/overthelex/SecondLayer}.

\bibliographystyle{plainnat}
\bibliography{references}

\end{document}